\documentclass[A4paper, 11 pt, conference]{ieeeconf}  
\IEEEoverridecommandlockouts       
\usepackage{geometry}
\usepackage{amsmath}
\usepackage{booktabs}
\usepackage{multirow}
\usepackage{tabularx}
\usepackage{array}
\usepackage{xcolor}
\usepackage{graphics} 
\usepackage{epsfig} 
\usepackage{mathptmx} 
\usepackage{times} 
\usepackage{amsmath} 
\usepackage{amssymb}  
\usepackage{subfigure}
\usepackage{algorithm} 
\usepackage{algorithmic} 
\usepackage{multirow} 
\usepackage{hyperref}
\hypersetup{hidelinks}
\begin{document}

\title{\LARGE \bf
Infrared and Visible Image Fusion Based on Implicit Neural Representations}

\author{Shuchen Sun, Ligen Shi, Chang Liu,  Lina Wu and  Jun Qiu$^*$ 
\thanks{
Shuchen Sun, Chang Liu, Lina Wu, and Jun Qiu are with the College of Computer Science, Beijing Information Science and Technology University, Beijing 100101, China .
Ligen Shi is with the School of Mathematical Sciences, Capital Normal University, Beijing 100048, China . 
Corresponding author: Jun Qiu.
}}
\maketitle 
\thispagestyle{empty}

\begin{abstract}
Infrared and visible light image fusion aims to combine the strengths of both modalities to generate images that are rich in information and fulfill visual or computational requirements. This paper proposes an image fusion method based on Implicit Neural Representations (INR), referred to as INRFuse. This method parameterizes a continuous function through a neural network to implicitly represent the multimodal information of the image, breaking through the traditional reliance on discrete pixels or explicit features. The normalized spatial coordinates of the infrared and visible light images serve as inputs, and multi-layer perceptrons is utilized to adaptively fuse the features of both modalities, resulting in the output of the fused image. By designing multiple loss functions, the method jointly optimizes the similarity between the fused image and the original images, effectively preserving the thermal radiation information of the infrared image while maintaining the texture details of the visible light image. Furthermore, the resolution-independent characteristic of INR allows for the direct fusion of images with varying resolutions and achieves super-resolution reconstruction through high-density coordinate queries. Experimental results indicate that INRFuse outperforms existing methods in both subjective visual quality and objective evaluation metrics, producing fused images with clear structures, natural details, and rich information without the necessity for a training dataset.
\end{abstract}

\section{Introduction}
Traditional image processing mainly relies on information from a single modality and struggles to effectively handle complex real-world scenarios. As a result, image fusion technology has become a key research direction for enhancing perception and decision-making capabilities.Infrared and visible light image fusion represents a fundamental aspect of image fusion technology. Researchers both domestically and internationally have undertaken extensive studies in this field, encompassing both traditional fusion techniques and those based on deep learning. Traditional methods~\cite{burt1987laplacian} depend on manually designed feature extraction rules, which often fail to adequately account for the differences between image modalities, thereby limiting the ability to harness their complementary information. Furthermore, these approaches exhibit limited adaptability and lack the capability to be adjusted according to varying scenarios or image characteristics.

The advantages of deep learning in feature extraction and data representation have led to the development of a growing number of deep learning-based fusion methods~\cite{liu2018infrared}. These methods excel in feature extraction and detail enhancement. However, due to their reliance on pixel-level explicit operations, they often face challenges in delicately processing image details, particularly in high-frequency regions under complex scenarios. Additionally, deep learning approaches typically rely on fixed network architectures and training strategies, which may not adaptively accommodate modality differences. Moreover, as image resolution increases, there is a corresponding rise in storage and computational overhead.

To address the limitations of existing methods, this paper proposes an end-to-end fusion method based on Implicit Neural Representation (INR). INR is a technique that utilizes neural networks to learn continuous representations of data~\cite{chen2021learning}, treating images as continuous implicit functions and generating their representations through neural network parameterization to achieve high visual quality.In the proposed method, the neural network learns a continuous mapping function to implicitly model the multimodal information of images, efficiently fusing the complementary characteristics of visible light and infrared images through a joint optimization strategy. Compared to traditional methods, INRFuse eliminates the need for manually designed fusion rules, automatically constructing the optimal fusion strategy through end-to-end learning. Additionally, its continuous representation avoid interpolation errors caused by discrete representations. Unlike deep learning methods that require large training datasets, INRFuse demonstrates high robustness with fewer network parameters and does not rely on extensive training data.

The main contributions of this paper can be summarized as follows:
\begin{itemize}
\item A novel image fusion method for infrared and visible light images based on INR is proposed. This method effectively processes data from different sensors through implicit neural representations, avoiding the detail loss caused by traditional discrete pixel representations and facilitating the precise capture of fine image structures.
\item A multi-constraint loss function incorporating pixel consistency, structural preservation, and sparsity regularization is designed to achieve unsupervised high-quality fusion.
\item This method offers a general framework for multimodal fusion, supporting the fusion of images with different resolutions as well as super-resolution reconstruction of the fused image.
\end{itemize}
\section{Method}
\begin{figure*}[htbp]
    \centering
    \includegraphics[width=\textwidth]{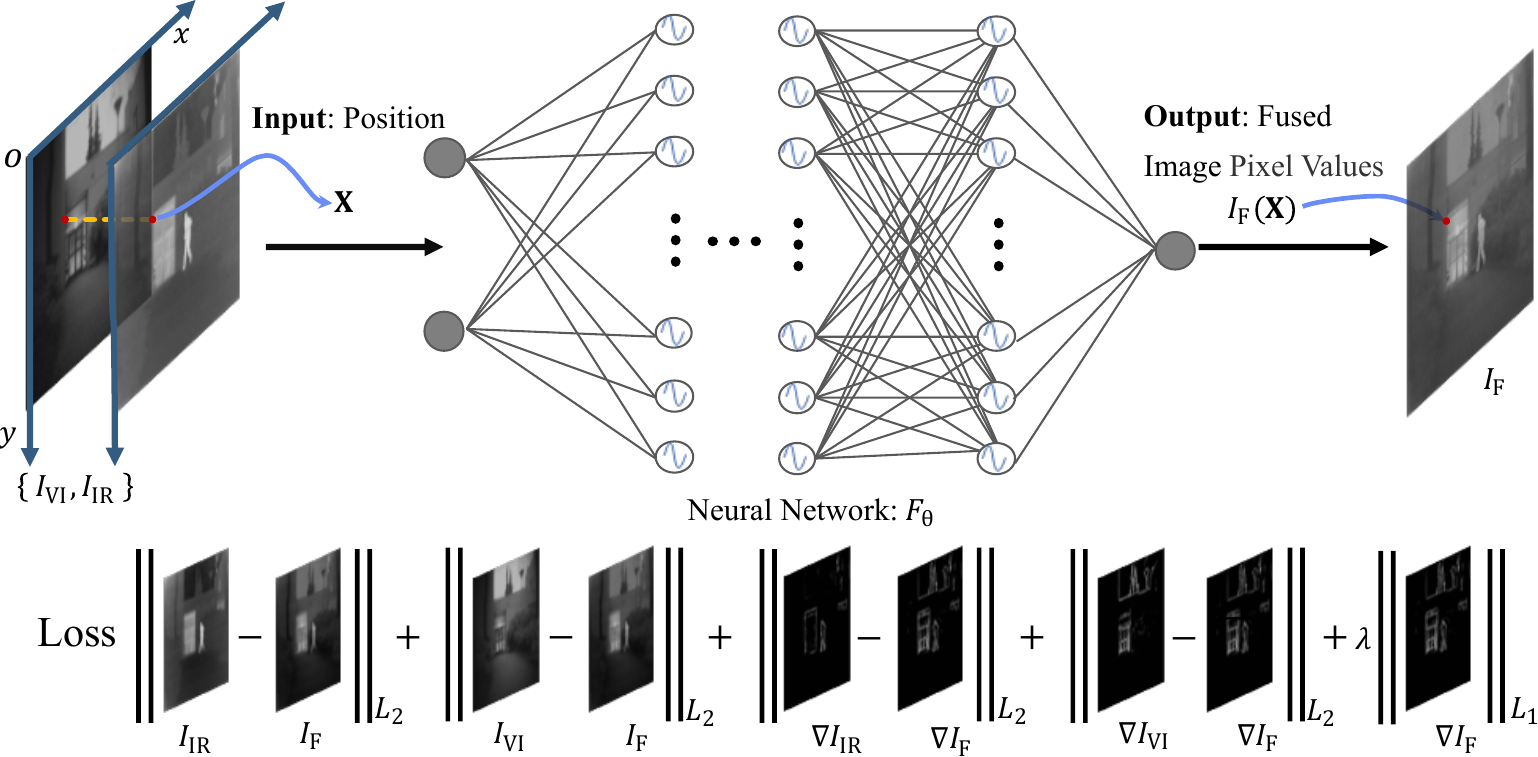} 
    \caption{Infrared-Visible Light Fusion Framework Based on Implicit Neural Representations.}
    \label{fig1}
\end{figure*}

\subsection{Parameterized Representation of Image Fusion}\label{AA}

This paper proposes an infrared and visible image fusion method based on Implicit Neural Representation (INR). The method constructs a continuous mapping function that maps spatial coordinates $\mathbf{X} = (x, y)$ to the pixel values of the fused image, as defined by:
\begin{equation}
F_{\theta}: \mathbf{X} \mapsto I_{\text{F}}(\mathbf{X}), \tag{1}
\end{equation}
where $I_{\text{F}}$ denotes the implicit function parameterized by a neural network $F_{\theta}$. The input is the image coordinate $\mathbf{X}$, and the output is the corresponding pixel value in the fused image.

By parameterizing this continuous function with a neural network, the model learns a joint implicit representation that simultaneously encodes information from the infrared image $I_{\text{IR}}(\mathbf{X})$ and the visible image $I_{\text{VIS}}(\mathbf{X})$. During the optimization process, the network adaptively learns the correspondence between the two modalities and coherently maps them into a unified representational space, thus enabling effective multi-modal fusion. The overall framework of the proposed fusion method is illustrated in Fig.~\ref{fig1}.

\subsection{Neural Network Model}
Sinusoidal Representation Networks (SIREN)~\cite{sitzmann2020implicit} enhance the modeling capacity for high-frequency information by incorporating the sine function as the activation function within Multilayer Perceptrons (MLPs). From a physical perspective, both visible and infrared light are electromagnetic waves whose intrinsic oscillatory nature can be naturally represented by sinusoidal functions. This inherent alignment between the physical characteristics of light and the mathematical form of sine functions makes SIREN particularly well-suited for modeling multi-modal image signals. To address the common issue of gradient vanishing associated with sine activations, researchers have developed a specialized parameter initialization scheme and demonstrated SIREN stability and reliability in fitting complex signals and their derivatives.

In this paper, we construct MLPs network consisting of 5 layers, each containing 256 neurons, with the sine function applied as the activation in all hidden layers:
\begin{equation}
\phi_i(\mathbf{X}_i) = \sin\left( \mathbf{W}_i \mathbf{X}_i + \mathbf{b}_i \right),\tag{2}
\end{equation}
where $\phi_i : \mathbb{R}^{M_i} \mapsto \mathbb{R}^{N_i}$ is the $i^{th}$ layer of the network. It consists of the affine transform defined by the weight matrix $\mathbf{W}_i \in \mathbb{R}^{N_i \times M_i}$ and the biases $\mathbf{b}_i \in \mathbb{R}^{N_i}$ applied on the input $\mathbf{X}_i \in \mathbb{R}^{M_i}$, followed by the sine nonlinearity applied to each component of the resulting vector.

\subsection{Loss Function}
The loss function design should consider both the salient targets of the infrared image and the texture details of the visible image. To achieve this, this paper proposes a loss function with three components: pixel consistency loss, gradient consistency loss, and regularization loss. The pixel consistency loss calculates the $ L_1 $ norm difference between the fused image and the source images to reduce distortion during the fusion process. The gradient consistency loss constrains the difference between the fused image and the source images in the gradient domain to preserve edge and structural information. The regularization term is used to suppress noise and enhance the smoothness of the fused image.The formulation is as follows:
{\footnotesize
\begin{equation}
\begin{aligned}
\mathcal{L} = & \, \| I_{\text{IR}}(\mathbf{X}) - I_{\text{F}}(\mathbf{X}) \|_2^2 + \| I_{\text{VIS}} (\mathbf{X})- I_{\text{F}}(\mathbf{X}) \|_2^2 \\
& + \| \nabla I_{\text{IR}}(\mathbf{X}) - \nabla I_{\text{F}}(\mathbf{X}) \|_2^2 + \| \nabla I_{\text{VIS}}(\mathbf{X}) - \nabla I_{\text{F}}(\mathbf{X}) \|_2^2 \\
& + \lambda \| \nabla I_{\text{F}}(\mathbf{X}) \|_1,
\end{aligned}
\tag{3}
\end{equation}}
where \( \lambda \) is the hyperparameter for sparse regularization loss, \( \nabla\) represents the gradient operation.
\section{Experiments and Analysis}
\subsection{Dataset and Experimental Details}
In this study, two mainstream datasets were primarily employed to validate the proposed INRFuse method: the TNO dataset~\cite{toet2014tno} and the RoadScene dataset~\cite{xu2020u2fusion}. 

In the experiments, the pixel values of the infrared and visible light images were normalized to the range of \([-1, 1]\) with a standard deviation of 0.5. The learning rate was set to 0.001, the optimizer utilized was Adam, and the hyperparameter \( \lambda \) was set to 1. All experiments were conducted on a computer equipped with an NVIDIA GeForce RTX 4060 Laptop GPU, and the deep learning models were implemented using the PyTorch framework version 2.4.1 with Python 3.8.20.
\subsection{Comparison Methods and Fusion Metrics}
To validate the effectiveness of the proposed INRFuse, we compared it with nine advanced infrared and visible image fusion algorithms. These include traditional methods such as CVT~\cite{nencini2007remote} and MDLatLRR~\cite{li2020mdlatlrr}. Two CNN-based fusion algorithms, DenseFuse~\cite{li2018densefuse} and U2Fusion~\cite{xu2020u2fusion}. Two GAN-based fusion algorithms, FusionGAN~\cite{ma2019fusiongan}, and UMFusion~\cite{wang2205unsupervised}. Three Transformer-based fusion algorithms, DATFuse~\cite{tang2023datfuse}, PPTFusion~\cite{fu2021ppt}, and YDTR~\cite{Tang_2023_YDTR}. 

To comprehensively and objectively evaluate the fused image, this paper employs six commonly used image fusion quantitative evaluation metrics: Entropy (EN)~\cite{roberts2008assessment}to measure the amount of information in the fused image. Standard Deviation (SD)~\cite{rao1997fibre}reflects the overall contrast of the fused image. Correlation Coefficient (CC)~\cite{ma2019infrared}measures the degree of linear correlation between the fused image and the source images. Sum of the Correlations of Differences (SCD)~\cite{aslantas2015new} assesses the extent to which information from the source images is preserved in the fused image. Multi-Scale Structural Similarity Index Measure (MS-SSIM)~\cite{ma2015perceptual}provides a more accurate prediction of image quality by analyzing the structural similarity between the fused image and the source images at different scales. Visual Information Fidelity (VIF) ~\cite{han2013new}measures the fidelity of information in the fused image.
\begin{figure}[htbp]
    \centering
\includegraphics[width=\linewidth,height=0.90\textheight,keepaspectratio]{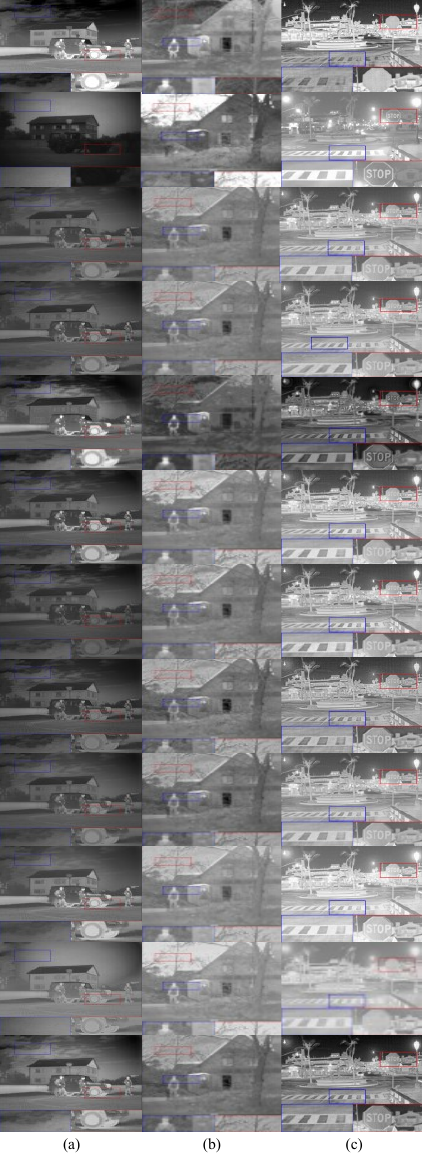} 
    \caption{(a) (b) represent the fusion results of different methods on the TNO dataset, and (c) represents the fusion results of different methods on the RoadScene dataset.From top to bottom: infrared image, visible image, the fusion results of CVT, MDLatLRR, FusionGAN, UMFusion, PPTFusion, U2Fusion, DenseFuse, YDTR, DATFuse, and the proposed INRFuse.}
    \label{fig2}
\end{figure}
\begin{table}[htbp]
    \centering
    \caption{Quantitative Evaluation Results of Nine Representative Infrared and Visible Light Image Fusion Methods and the Proposed INRFuse on the TNO Dataset. }
    \renewcommand{\arraystretch}{1.2}
    \setlength{\tabcolsep}{4pt}
    \begin{tabular}{
        >{\centering\arraybackslash}p{1.5cm}   
        >{\centering\arraybackslash}p{0.7cm}   
        >{\centering\arraybackslash}p{0.7cm}   
        >{\centering\arraybackslash}p{0.7cm}   
        >{\centering\arraybackslash}p{0.7cm}   
        >{\centering\arraybackslash}p{0.7cm}   
        >{\centering\arraybackslash}p{1.3cm}   
    }
        \toprule
        \textbf{Method} & \textbf{SD↑} & \textbf{VIF↑} & \textbf{CC↑} & \textbf{SCD↑} & \textbf{EN↑} & \textbf{MS\_SSIM↑} \\
        \midrule
        FusionGAN  & 8.6594 & 0.5672 & 0.4233 & 1.2989 & 6.3959 & 0.7167 \\
        UMFusion   & 9.1662 & 0.7032 & 0.5264 & \textbf{\textcolor{cyan}{1.6623}} & \textbf{\textcolor{cyan}{6.7367}} & 0.8920 \\
        PPTFusion  & 8.9790 & 0.6615 & 0.5346 & 1.5774 & 6.5779 & 0.8567 \\
        U2Fusion   & 9.3171 & 0.6603 & 0.5392 & 1.5893 & 6.5924 & \textbf{\textcolor{cyan}{0.9025}} \\
        DenseFuse  & 9.0120 & 0.6489 & \textbf{\textcolor{red}{0.5495}} & 1.5796 & 6.5189 & 0.8665 \\
        YDTR       & \textbf{\textcolor{cyan}{9.3768}} & 0.7179 & 0.4978 & 1.5500 & 6.6549 & 0.8356 \\
        DATFuse    & 9.1699 & \textbf{\textcolor{cyan}{0.7525}} & 0.4879 & 1.4484 & 6.6032 & 0.7974 \\
        \midrule
        CVT        & 9.0793 & 0.6489 & 0.5463 & 1.5844 & 6.5665 & 0.8690 \\
        MDLatLRR   & 9.0651 & 0.6825 & 0.5472 & 1.5982 & 6.5583 & 0.8911 \\
        Ours       & \textbf{\textcolor{red}{9.5855}} & \textbf{\textcolor{red}{0.7609}} & \textbf{\textcolor{cyan}{0.5485}} & \textbf{\textcolor{red}{1.8014}} & \textbf{\textcolor{red}{6.9986}} & \textbf{\textcolor{red}{0.9183}} \\
        \bottomrule
    \end{tabular}
    \label{table1}
\end{table}
\begin{table}[htbp]
    \centering
    \caption{Quantitative evaluation results of nine representative infrared and visible image fusion methods and the proposed INRFuse on the Roadscene dataset.}
    \renewcommand{\arraystretch}{1.2}
    \setlength{\tabcolsep}{4pt}
   \begin{tabular}{
        >{\centering\arraybackslash}p{1.5cm}   
        >{\centering\arraybackslash}p{0.7cm}   
        >{\centering\arraybackslash}p{0.7cm}   
        >{\centering\arraybackslash}p{0.7cm}   
        >{\centering\arraybackslash}p{0.7cm}   
        >{\centering\arraybackslash}p{0.7cm}   
        >{\centering\arraybackslash}p{1.3cm}   
    }
        \toprule
        \textbf{Method} & \textbf{SD↑} & \textbf{VIF↑} & \textbf{CC↑} & \textbf{SCD↑} & \textbf{EN↑} & \textbf{MS\_SSIM↑} \\
        \midrule
        FusionGAN  & 9.4069 & 0.5774 & 0.6194 & 0.7775 & 6.5846 & 0.7546 \\
        UMFusion   & 9.9195 & \textbf{\textcolor{cyan}{0.7992}} & 0.7441 & \textbf{\textcolor{cyan}{1.2881}} & \textbf{\textcolor{cyan}{7.1087}} & \textbf{\textcolor{cyan}{0.9156}} \\
        PPTFusion  & 9.6836 & 0.7539 & 0.7526 & 1.2299 & 6.9991 & 0.8895 \\
        U2Fusion   & 10.037 & 0.7376 & 0.7411 & 1.0805 & 6.9640 & 0.9141 \\
        DenseFuse  & 9.7479 & 0.7492 & \textbf{\textcolor{red}{0.7621}} & 1.1707 & 6.9162 & 0.8880 \\
        YDTR       & \textbf{\textcolor{red}{10.162}} & 0.7915 & 0.7370 & 1.1788 & 6.9651 & 0.8862 \\
        DATFuse    & 9.9428 & 0.7969 & 0.7183 & 0.9030 & 6.8263 & 0.8125 \\
        \midrule
        CVT        & 9.7623 & 0.7492 & 0.7604 & 1.1727 & 6.9427 & 0.8900 \\
        MDLatLRR   & 9.7703 & 0.7827 & 0.7596 & 1.1903 & 6.9502 & 0.9151 \\
        Ours       & \textbf{\textcolor{cyan}{10.072}} & \textbf{\textcolor{red}{0.8180}} & \textbf{\textcolor{cyan}{0.7619}} & \textbf{\textcolor{red}{1.6173}} & \textbf{\textcolor{red}{7.2728}} & \textbf{\textcolor{red}{0.9361}} \\
        \bottomrule
    \end{tabular}
    \label{table2}
\end{table}
\subsection{Experimental Results and Discussion}
\begin{figure*}[htbp]
    \centering
    \includegraphics[scale=0.6]{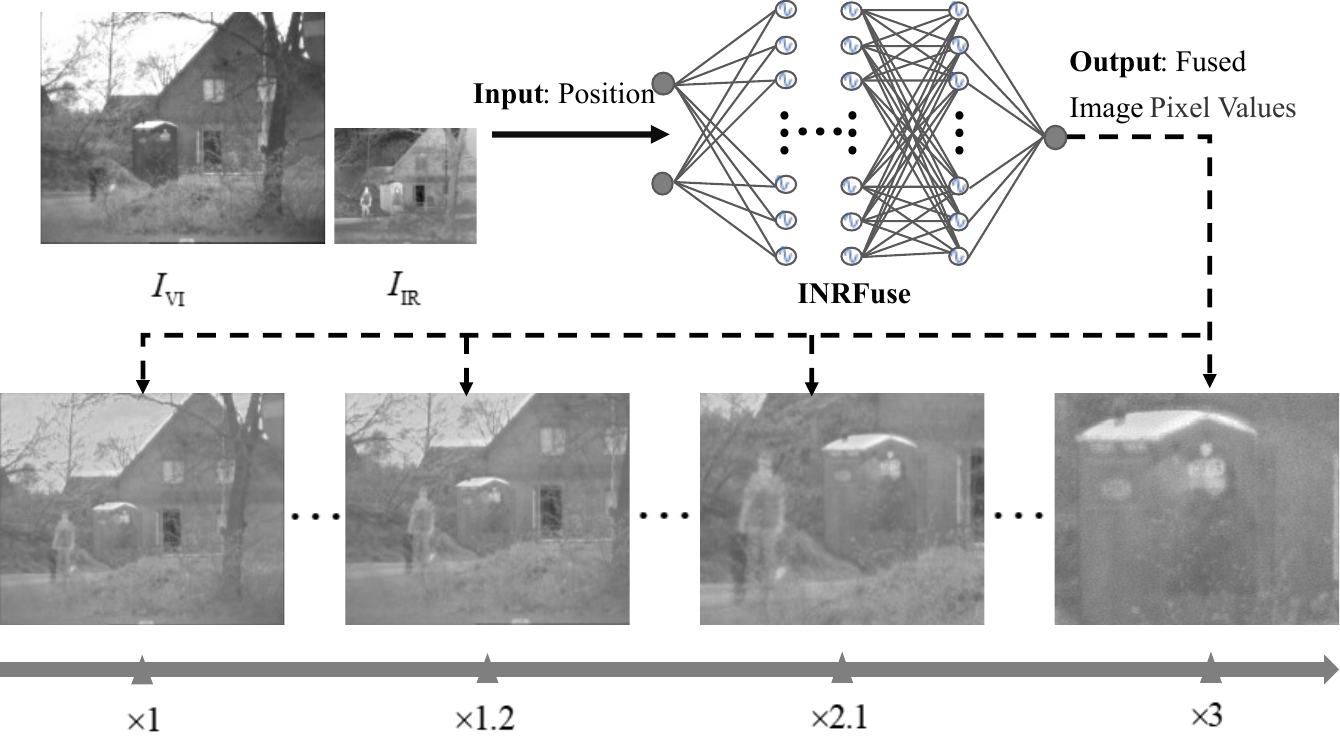} 
    \caption{Fusion of Images with Varying Resolutions and Super-Resolution Reconstruction of Fused Images.}
    \label{fig3}
\end{figure*}
\paragraph{TNO Dataset}
To validate the effectiveness of the proposed method, experiments were conducted on 20 pairs of images from the TNO dataset. Two representative infrared and visible images were selected for subjective evaluation. As shown in Fig. \ref{fig2} (a) and (b) respectively show two pairs of source images from the TNO dataset and their corresponding fused images obtained through ten different methods.
All ten methods achieved satisfactory imaging results. 
Specifically, traditional methods such as CVT and MDLatLRR employ fixed transformation strategies that cannot adaptively adjust feature extraction based on input images, resulting in brightness loss in the fused images. FusionGAN, which incorporates an adversarial learning mechanism, preserves sharpened infrared targets, but the visible details suffer from blurring and loss. For example, as shown in (b), the leaves exhibit a blurred result. PPTFusion, DenseFuse, and U2Fusion exhibit weaker capability in extracting infrared thermal radiation information, resulting in lower contrast in the fused images, as seen in the tire area of (a). YDTR and UMFusion generally achieve better fusion results, but exhibit some blurring in the details compared to the proposed INRFuse, such as in the roof area of (a). DATFuse fails to effectively preserve infrared thermal radiation information, leading to the loss of infrared details, as observed in the sky area of (a) and the leaves in (b). 

Experiments were conducted using ten different algorithms on 20 image pairs from the TNO dataset, and their performance was evaluated using six objective metrics, as presented in Tab. \ref{table1}. In these metrics, an upward arrow (↑) indicates that a higher score corresponds to better image fusion quality. For each evaluation criterion, the best and second-best methods are highlighted in bold red and bold cyan, respectively. The results demonstrate that the proposed INRFuse method achieves the best performance in five key metrics: SD, VIF, SCD, EN, and MS-SSIM, outperforming other existing methods. Although INRFuse does not achieve absolute optimal performance in the CC metric, the gap between its performance and the best-performing method is minimal. 
\paragraph{Roadscene Dataset}

In this section, we conducted further experimental validation of the proposed INRFuse on the RoadScene dataset using 40 pairs of infrared and visible light images. Additionally,  we selected a representative pair of infrared and visible light images for subjective evaluation, as shown in Fig. 2 (c). Compared to the other nine methods, the fusion images produced by the proposed INRFuse exhibit higher overall contrast, better highlighting typical thermal radiation information such as pedestrian targets. For typical details such as signboard in (c), the fusion results by the proposed INRFuse are complete and clear.

As shown in Tab. \ref{table2}, it can be observed that the proposed INRFuse achieved the best scores in four metrics: VIF, SCD, EN, and MS-SSIM, outperforming other existing methods. Although INRFuse did not achieve the absolute best performance in the CC and SD metrics, its performance was the second best among the ten methods.

\subsection{Multi-Resolution Image Fusion and Super-Resolution of the Fused Image}
Imaging systems across different modalities often operate at distinct spatial resolutions. Traditional approaches typically necessitate upsampling, downsampling, or alignment preprocessing prior to fusion, which can lead to interpolation errors and loss of information. INR model images as continuous functions that map normalized spatial coordinates to pixel values, facilitating a resolution-independent representation. By decoupling the image signal from the discrete pixel grid, both low- and high-resolution images can be jointly optimized under the same functional framework through coordinate points with varying densities, thus facilitating cross-resolution fusion. During training, to ensure the comparability of loss terms between images of different resolutions, the fused image is spatially aligned with the source images before loss computation, thereby ensuring the stability and accuracy of the fusion process.

Furthermore, the continuous spatial representation provided by INR allows for querying higher-density coordinates during inference, enabling detailed super-resolution reconstruction. The proposed INRFuse additionally leverages the high-frequency sensing capabilities of the SIREN network, effectively capturing multi-scale structures and texture information across modalities. This presents a unified and high-fidelity modeling paradigm for the fusion of infrared and visible light images. As shown in Fig. \ref{fig3}, the fusion results of infrared and visible light images at different resolutions are displayed, with super-resolution reconstructions randomly performed at ×1.2, ×2.1, and ×3, thereby validating the effectiveness of the proposed method in cross-resolution fusion and super-resolution reconstruction.

\section{Conclusions}
This paper proposes a novel visible-infrared image fusion method, INRFuse, based on INR. INRFuse models multi-modal image information as a continuous function parameterized by a neural network, enabling adaptive fusion through joint optimization. By learning the correspondence between visible and infrared features, the method effectively integrates complementary information from both modalities. Unlike traditional pixel- or feature-level fusion approaches, INRFuse removes the reliance on discrete pixel grids, allowing for smoother integration and improved detail fidelity. It also supports resolution-independent representation, enabling direct fusion across varying resolutions and super-resolution reconstruction via high-density coordinate querying. INRFuse operates on a single image pair without requiring large-scale training data, demonstrating strong adaptability and practical value for real-world multi-modal fusion tasks.
\section*{Acknowledgment}
This work was supported by National Natural Science Foundation of China (under Grant Nos. 62171044, 61931003).

%
%




\end{document}